\documentclass{article}
\usepackage{spconf,amsmath,graphicx}
\usepackage{booktabs}
\usepackage{multirow}
\usepackage{xcolor}
\usepackage{amsfonts}
\usepackage{subcaption}
\usepackage{pifont}
\usepackage{algorithm} 
\usepackage{algorithmic}
\usepackage[most]{tcolorbox}

\usepackage{etoolbox} 

\makeatletter
\patchcmd{\thebibliography}{\settowidth}{\setlength{\itemsep}{0ex} \settowidth}{}{}
\makeatother

\usepackage[breaklinks=true,bookmarks=false]{hyperref}

\title{Audio Visual Segmentation Through Text Embeddings}
\name{Kyungbok Lee$^{1}$, You Zhang$^{2}$, Zhiyao Duan$^{2}$}
\address{
    $^{1}$Department of Computer Science, University of Rochester, Rochester, NY, USA \\
    $^{2}$Department of Electrical and Computer Engineering, University of Rochester, Rochester, NY, USA \\
    \texttt{klee109@u.rochester.edu}, \quad \texttt{\{you.zhang, zhiyao.duan\}@rochester.edu}
}

\begin{document}
%
\maketitle
\begin{abstract}
The goal of Audio-Visual Segmentation (AVS) is to localize and segment the sounding source objects from video frames. Research on AVS suffers from data scarcity due to the high cost of fine-grained manual annotations. Recent works attempt to overcome the challenge of limited data by leveraging the vision foundation model, Segment Anything Model (SAM), prompting it with audio to enhance its ability to segment sounding source objects. While this approach alleviates the model's burden on understanding visual modality by utilizing knowledge of pre-trained SAM, it does not address the fundamental challenge of learning audio-visual correspondence with limited data. To address this limitation, we propose \textbf{AV2T-SAM}, a novel framework that bridges audio features with the text embedding space of pre-trained text-prompted SAM. 
Our method leverages multimodal correspondence learned from rich text-image paired datasets to enhance audio-visual alignment. Furthermore, we introduce a novel feature, $\mathbf{\textit{\textbf{f}}_{CLIP} \odot \textit{\textbf{f}}_{CLAP}}$, which emphasizes shared semantics of audio and visual modalities while filtering irrelevant noise. Our approach outperforms existing methods on the AVSBench dataset by effectively utilizing pre-trained segmentation models and cross-modal semantic alignment. The source code is released at \textcolor{magenta}{\url{https://github.com/bok-bok/AV2T-SAM}}.
\end{abstract}

\section{Introduction}
\label{sec:intro}

\begin{figure*}[h!]
  \centering
  \includegraphics[width=\textwidth]{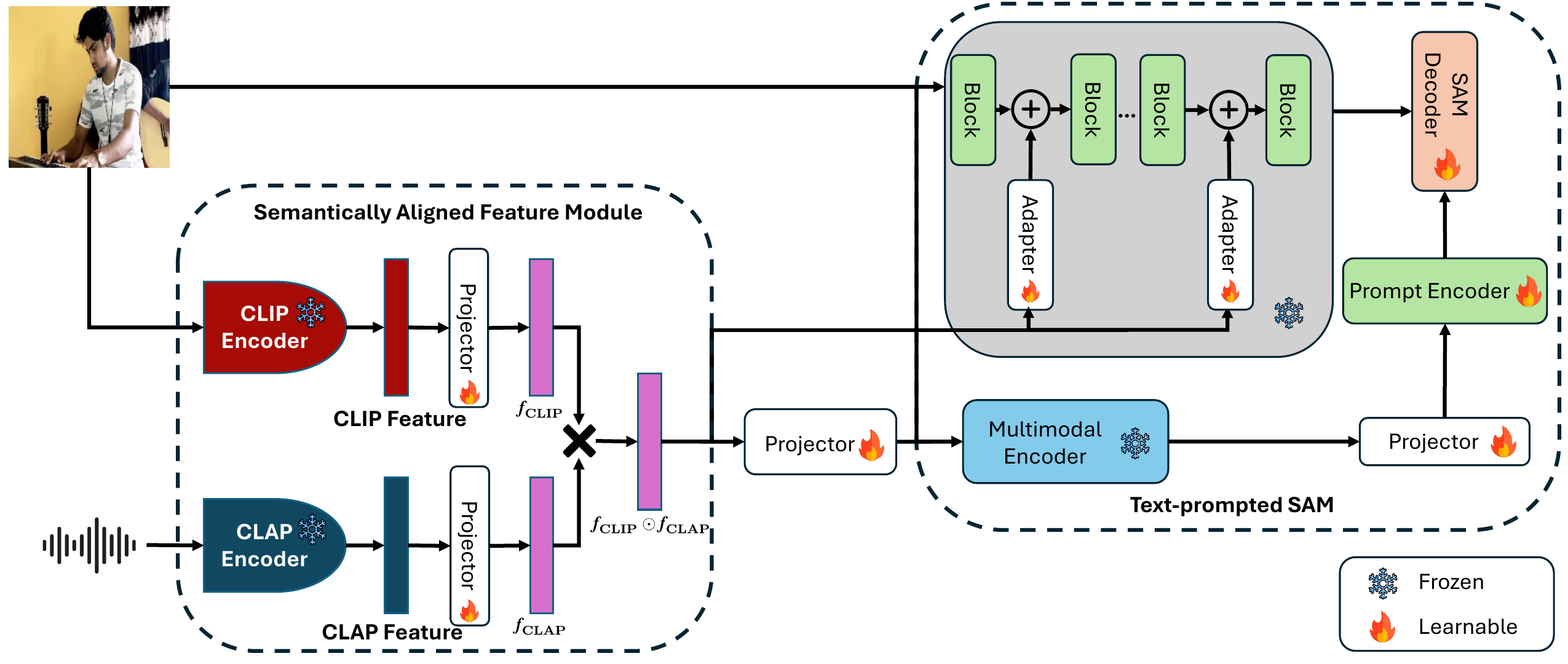}
  \caption{The overview of our proposed AV2T-SAM framework.  Algorithm~\ref{algorithm} specifies the process of the AV2T-SAM framework.}
  \label{fig:model_structure}
  \vspace{-12pt}
\end{figure*}

Audio-visual segmentation (AVS) is the task of segmenting sounding source objects from the video frames. Without pixel-level supervision, early researchers tackled the sounding source localization (SSL) task by leveraging self-supervised learning~\cite{chen2021sslhard, arandjelovic2018sslobjects}
, exploiting the semantic alignment between image and audio pair data. However, this approach results in coarse localization of sounding objects, which limits the application in fields that require fine-grained segmentation masks. Zhou \textit{et al.}~\cite{zhou2022avs} introduced the AVSBench dataset, annotated with a pixel-level segmentation map. The initial works on this dataset often fuse audio and image embeddings~\cite{zhou2022avs, gao2024avsegformer} and then decode the segmentation mask from the fused audio-visual embeddings. 

The Segment Anything Model (SAM)~\cite{kirillov2023sam, ravi2024sam2} introduces an innovative approach to segmentation tasks by leveraging large-scale pre-training, similar to the way large language models use prompt-based learning~\cite{liu2023prompting}. The prompting idea has been applied to AVS.
Mo and Tian ~\cite{mo2023av} pioneered the idea of audio-prompted SAM by providing audio as a prompt to the SAM decoder. Liu \textit{et al.}~\cite{liu2024annotation} advanced the audio-prompted SAM by introducing adapters into the image encoder of SAM. Audio-prompted SAM achieved better performance on the AVS task by using the image segmentation foundation model, which mitigates the problem of the limited dataset. However, while SAM empowers the capacity of image understanding, the burden of learning audio-visual correspondence persists due to the scarcity of labeled data.


To address the data scarcity issue, we propose \textbf{AV2T-SAM}, which projects audio prompts into a text embedding space. This enables the use of knowledge from pre-trained text-prompted SAM models. Text-image models benefit from a wealth of text-image paired datasets, surpassing the availability of audio-visual data. By transforming audio inputs into a format compatible with pre-trained text-prompted SAM, AV2T-SAM not only capitalizes on SAM’s segmentation expertise but also enhances the learning of audio-visual correspondence through access to the text embedding space.

Specifically, we introduce a novel feature, $\mathbf{\textit{\textbf{f}}_{CLIP} \odot \textit{\textbf{f}}_{CLAP}}$, which aligns audio and visual embeddings into a shared semantic space, leveraging the pre-trained cross-modal encoders CLIP~\cite{radford2021clip} and CLAP~\cite{elizalde2023clap}. By leveraging modality-specific projections, the model combines these embeddings through element-wise multiplication, capturing the intersection of audio and visual modalities. This design emphasizes shared semantics while reducing irrelevant information, producing a more robust and contextually enriched representation. Through this approach, our model effectively enhances audio-visual correspondence, enabling more accurate and reliable segmentation performance. We also identify a vision bias issue in the single sound source dataset (S4) of AVSBench, as our proposed system outperformed previous state-of-the-art models without using any audio information. 

In summary, our contributions are as follows: 
\begin{itemize}
    \item We propose \textbf{AV2T-SAM} (Audio-Visual to Text SAM), a novel framework of utilizing a pre-trained text-prompted SAM by projecting audio and visual features into a text embedding space. This strategy enables us to effectively harness the rich semantic information embedded in large-scale pre-trained cross-modal encoders, addressing the limitations posed by the scarcity of audio-visual segmentation masks.
    Experimental results demonstrate the effectiveness of our method. 
    \item We introduce a $\mathbf{\textit{\textbf{f}}_{CLIP} \odot \textit{\textbf{f}}_{CLAP}}$ feature that captures the intersection of audio and visual modalities, emphasizing shared semantics and filtering out irrelevant information, resulting in a more contextually enriched and discriminative feature for segmentation tasks.
    \item We demonstrate a serious vision bias in the AVSBench S4 dataset by surpassing previous state-of-the-art approaches without using any audio information.
\end{itemize}

\section{Related works}
\subsection{Multimodal Learning with Segment Anything Model}
Segment Anything Model (SAM)~\cite{kirillov2023sam} is the first foundation model for image segmentation tasks with various prompts. The model is pretrained with more than 11M images and 1B masks. Many studies verified capability of SAM on diverse downstream tasks, including medical image segmentation~\cite{ma2024medisam} and pose estimation~\cite{lin2024posesam}.

Beyond vision tasks, SAM is also widely used in multimodal learning due to its flexible prompt interactions. Wang et al.\cite{wang2024remotesam} employ remote sensing data to prompt SAM, while Liu et al.\cite{liu2024annotation} use audio to segment sounding objects. Researchers have also explored text-based prompting for SAM~\cite{zhang2024evfsam}.
In this study, we introduce applying a pretrained text-prompted SAM on audio-visual segmentation.

\subsection{Audio-Visual Segmentation}
With the trends of multimodal learning, audio-visual tasks also received much attention. Early researchers worked on the sound source localization task \cite{Zhao_2018_pixel,Chen_2021_hard} that aims to localize the sounding object in a frame. They often turned to self-supervised learning due to the lack of labels for localization. Though being effective to some extent, its application to fine-grained tasks like segmentation remains limited.

Zhou \textit{et al.}~\cite{zhou2022avs} introduce Audio-Visual Segmentation (AVS) with a labeled AVSBench dataset containing fine-grained segmentation masks. With the new dataset, researchers began exploring audio-visual segmentation with supervised learning~\cite{zhou2022avs,li2023catr}. Gao \textit{et al.}~\cite{gao2024avsegformer} proposed a transformer-based architecture for the AVS task. Liu \textit{et al.}~\cite{liu2023avstwostages} introduced a framework that separates the segmentation process and the process of verifying sounding object. 

Mo and Tian ~\cite{mo2023av} first explored utilizing the segment anything model with an audio prompt. Claiming the two-layer SAM decoder, which was previously prompted, is too shallow to model the audio-visual correspondence, Liu \textit{et al.} ~\cite{liu2024annotation} introduced the adapter that introjects audio information to the frozen SAM encoder. Seon \textit{et al.} \cite{seon2024extending} employed temporal information which was not considered on sam-based models. 
Nguyen and Park~\cite{nguyen2024save} advanced the adapter mechanism by introducing an additional trainable layer to each block of the SAM encoder and utilized the residual audio encoder. In this work, we will exploit the text embedding space and the power of pretrained text-prompted SAM~\cite{zhang2024evfsam} to advance the performance of SAM~\cite{kirillov2023sam} on the AVS.

\section{Method}
\subsection{Problem Formation}
The input video of audio-visual segmentation (AVS) comprises of a series of visual frames and corresponding audio. The image frames are represented as $\mathbf{V} = \{\mathbf{v}_i\}^T_{i=1}$,  where ($\mathbf{v}_i \in \mathbb{R}^{3 \times H_i \times W_i}$)
where \( \mathbf{v}_i \) denotes the \( i\)-th visual frame with spatial dimensions \( H_i \times W_i \), and \( T \) is the total number of frames in the video. Each frame \( \mathbf{v}_i \) is paired with a corresponding 1-second audio waveform $\mathbf{a}_i \in \mathbb{R}^S$, where \( S \) represents the number of audio samples in 1 second.
The goal of the AVS task is to segment all sound source objects from each frame. The ground truth segmentation map is provided by binary masks 
$\mathbf{M} = \{\mathbf{m}_i\}_{i=1}^T$ where $\mathbf{m}_i \in \{0, 1\}^{H_i \times W_i}$

\subsection{Text-prompted Segmentation Module}
The pre-trained text-prompted SAM model is used in our framework as a backbone to leverage its robust text-visual correspondence knowledge, which is from
extensive text-image datasets~\cite{kazemzadeh2014referitgamedata, yu2016refcocodata,nagaraja2016refcocogdata}. While this module is interchangeable, we particularly used EVF-SAM \cite{zhang2024evfsam}. The model employs the multimodal encoder, which is more efficient than adopting the large language model and performs better than the simple text encoder, such as CLIP~\cite{radford2021clip}. While the original EVF-SAM model uses text as input, our method does not rely on explicit text. Instead, we map audio-visual features into the language space. To achieve this, we adapt their methods by removing the text encoder but keeping the multimodal encoder fixed, leveraging its strong adaptability to prompt SAM with text embeddings effectively.


\subsection{Semantically Aligned Feature Module}
To enhance the semantic alignment between audio and visual modalities, we propose $\mathbf{\textit{\textbf{f}}_{CLIP} \odot \textit{\textbf{f}}_{CLAP}}$ feature. Let $\mathbf{a} \in \mathbb{R}^{d_a}$ and $\mathbf{v} \in \mathbb{R}^{d_v}$ denote the normalized embeddings for audio and visual inputs, respectively, obtained from CLIP~\cite{radford2021clip} and CLAP~\cite{elizalde2023clap} encoders. These embeddings are projected into a shared embedding space of dimension $d_s$ using learned projection matrices $\mathbf{W}_a \in \mathbb{R}^{d_s \times d_a}$ and $\mathbf{W}_v \in \mathbb{R}^{d_s \times d_v}$. The projected embeddings are computed as follows:
\[
\mathbf{\textit{\textbf{f}}_{CLAP}} = \mathbf{W}_a \mathbf{a}, \quad \mathbf{\textit{\textbf{f}}_{CLIP}} = \mathbf{W}_v \mathbf{v},
\]
To capture the intersection of semantic information from both modalities, we calculate:
$$
\mathbf{\textit{\textbf{f}}_{CLIP} \odot \textit{\textbf{f}}_{CLAP}},
$$ 
where $\odot$ denotes the Hadamard (element-wise) product.

This operation ensures that the resulting feature emphasizes shared semantics between the two modalities while filtering out modality-specific noise. 



Then, we pass the feature $\mathbf{\textit{\textbf{f}}_{CLIP} \odot \textit{\textbf{f}}_{CLAP}}$ to the projection module, which consists of two MLP layers, to connect the features to the text embedding space required to prompt the pre-trained text-prompted segmentation model.


\subsection{Adapters}
Fine-tuning the image encoder of SAM requires considerable computing resources.
To efficiently fuse audio and visual information while not entirely fine-tuning the image encoder, Liu \textit{et al.}~\cite{liu2024annotation} suggested an audio adapter mechanism that trains only the adapter module which injects audio information to the frozen image encoder.

In this work, we employ the adapter module in the image encoder of SAM with $\mathbf{\textit{\textbf{f}}_{CLIP} \odot \textit{\textbf{f}}_{CLAP}}$ to effectively fuse prompt information to the image encoder. Specifically, for the $j$-th layer of a transformer block, the $j$-th adapter projects the prompt information to the output of the transformer block and repeats the prompt information along the spatial dimensions to match the dimensions of the image features, denoted as $\textbf{P}_j \in \mathbb{R}^{T \times H_iW_i \times C}$. The adapted features are then added to the output of the previous encoder layer, \(E_{j-1}\), to form the input to the \(j\)-th layer, \(\mathbf{X}_j\), as follows:
\begin{equation}
    \mathbf{X}_j = E_{j-1}(\mathbf{X}_{j-1}) + \mathbf{P}_j.
\end{equation}
This approach ensures the efficient fusion of audio and visual information while preserving computational efficiency by freezing the image encoder during training.

\begin{algorithm}[t]
\caption{AV2T-SAM Inference Pipeline}
\begin{algorithmic}[1]
\label{algorithm}
\REQUIRE Visual frames $\mathbf{V} = \{\mathbf{v}_i\}_{i=1}^T$, audio waveform $\mathbf{A} = \{\mathbf{a}_i\}_{i=1}^T$
\ENSURE Segmentation masks $\mathbf{M} = \{\mathbf{m}_i\}_{i=1}^T$
\FOR{each frame $\mathbf{v}_i$ and audio $\mathbf{a}_i$}
    \STATE Extract CLIP feature $\mathbf{v}_i \gets \text{CLIP}(\mathbf{v}_i)$
    \STATE Extract CLAP feature $\mathbf{a}_i \gets \text{CLAP}(\mathbf{a}_i)$
    \STATE Project features: $\mathbf{\textit{\textbf{f}}_{CLIP}}_i \gets \mathbf{W}_v \mathbf{v}_i$,

    \quad\quad\quad\quad\quad\quad\quad$\mathbf{\textit{\textbf{f}}_{CLAP}}_i \gets \mathbf{W}_a \mathbf{a}_i$
    \STATE Fuse features: $\mathbf{\textit{\textbf{f}}_{CLIP} \odot \textit{\textbf{f}}_{CLAP}}_i$
    \STATE Map to text embedding space via projection MLP
    \STATE Prompt the text-prompted SAM with projected $\mathbf{\textit{\textbf{f}}_{CLIP} \odot \textit{\textbf{f}}_{CLAP}}_i$
    \STATE Decode mask $\mathbf{m}_i$ with text-prompted SAM
\ENDFOR
\RETURN $\mathbf{M}$
\end{algorithmic}
\end{algorithm}

\subsection{Learning Objectives}
For training the model, we combined the Binary Cross-Entropy (BCE) loss and the Intersection over Union (IoU) loss, following the Liu \textit{et al.}~\cite{liu2024annotation}, 
\begin{equation}
    L_{\text{total}} = L_{\text{BCE}} + L_{\text{IoU}}.
\end{equation}

\section{Experimental Setup}
\subsection{Dataset}

AVSBench~\cite{zhou2022avs} is the pixel-level audio-visual segmentation dataset, which consists of 5-second videos downloaded from YouTube. AVSBench contains two datasets: Single Sound Source Dataset (S4) and Multi Sound Source Dataset (MS3).

\textbf{Single Sound Source dataset (S4).}
This dataset includes videos with only a single sound source. The dataset contains 4,932 videos in total over 23 categories and is split into training, validation, and test sets with 3,452, 740, and 740 samples, respectively. 
Notably, annotation details vary across subsets: only the first frame of each training video is annotated with binary masks, while the validation and test sets include full annotations for all five frames of every video.

\textbf{Multi Sound Sources dataset (MS3).}
The dataset contains 424 videos and is split into training, validation, and test sets with 296, 64, and 64 samples, respectively. For all splits, including training, every frame is annotated with the segmentation mask.

\subsection{Evaluation Metrics}
Following the original work~\cite{zhou2022avs}, we use the mean Intersection over Union ($M_{\mathcal{J}}$) and F-score ($M_{\mathcal{F}}$) to evaluate our model. The $M_{\mathcal{J}}$ computes the average IoU of predicted mask and ground truth masks over total frames, and $M_{\mathcal{F}}$ computes the harmonic mean of precision and recall.

\subsection{Implementation Details}
We utilize EVF-SAM2~\cite{zhang2024evfsam}, which integrates SAM-2-L~\cite{ravi2024sam2} as the SAM backbone and Beit-3-L~\cite{wang2022beit3} as the multimodal encoder for receiving a text prompt. For experiments using the SAM1~\cite{kirillov2023sam} as a backbone, we employ EVF-SAM1, which consists of SAM-H and Beit-3-L. The input image resolution was set to $1024 \times 1024$, and We train the model for 40 epochs using the AdamW optimizer.

\begin{figure*}
    \centering
    \includegraphics[width=\textwidth]{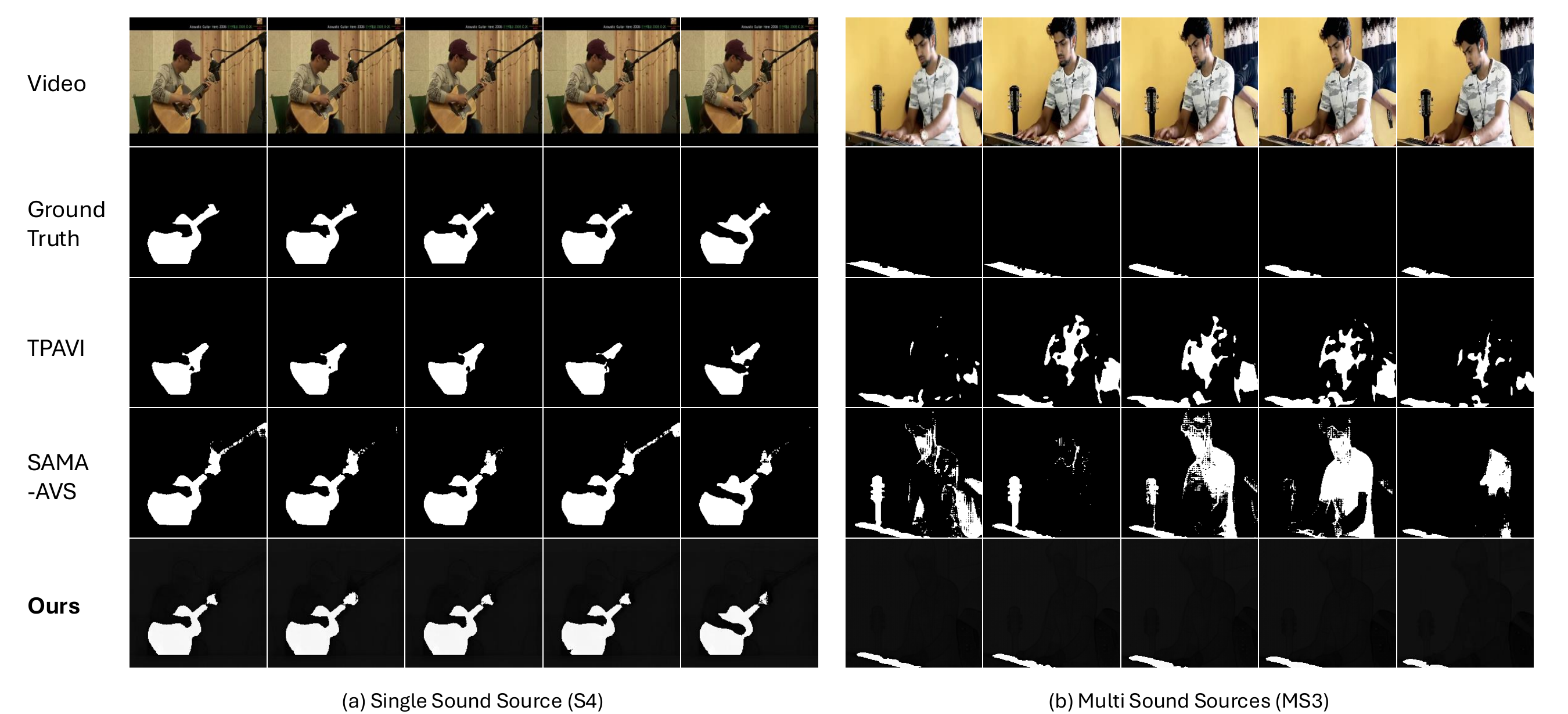}
    \caption{Comparison with other methods on examples. Our method successfully separates and segments a whole object compared to other methods.}
    \label{fig:examples}
    \vspace{-12pt}
\end{figure*}
\section{Experiments}
\subsection{Main Results}

\begin{table}[]

\centering
\scalebox{0.8}{
\begin{tabular}{llccccc}
\toprule
\textbf{Category} & \textbf{Method}& \textbf{Adapter} & \multicolumn{2}{c}{\textbf{S4 (V1S)}} & \multicolumn{2}{c}{\textbf{MS3 (V1M)}} \\ 
\cmidrule(lr){4-5} \cmidrule(lr){6-7}
 & & & $\mathcal{M}_{\mathcal{J}}$ & $\mathcal{M}_{\mathcal{F}}$ & $\mathcal{M}_{\mathcal{J}}$ & $\mathcal{M}_{\mathcal{F}}$ \\ 
\midrule
\multirow{8}{*}{\textbf{AVS}} 

 & AVSBench&- & 78.74 & 0.879 & 54.00 & 0.645 \\
 & ECMVAE&- & 81.74 & 0.901 & 57.84 & 0.708 \\
 & AVSegFormer&- & 82.06 & 0.899 & 58.36 & 0.693 \\
 & AVSC&- & 81.29 & 0.886 & 59.50 & 0.657 \\
 & AuTR&- & 80.40 & 0.891 & 56.20 & 0.672 \\
 & GAVS&- & 80.06 & 0.902 & 63.70 & 0.774 \\
 & AQFormer&- & 81.60 & 0.894 & 61.10 & 0.721 \\
 & COMBO&- & 84.70 & 0.919 & 59.20 & 0.712 \\ 
\midrule
 \multirow{7}{*}{\textbf{SAM}} 
 & AV-SAM& \ding{55} & 40.47 & 0.566 & - & - \\
 & AP-SAM& \ding{55} & 69.61 & 0.796 & 51.58 & 0.578 \\  
 & SAMA-AVS&\ding{51} & 81.53 & 0.886 & 63.14 & 0.691 \\
 & SAVE&\ding{51} & 85.11 & 0.912 & 67.01 & 0.739 \\
 & ST-BAVA&\ding{51} & 82.46&  0.906&  69.01&  0.776 \\
 & \textbf{AV2T-SAM} &\ding{55} &83.11 & 0.901 & 55.81 & 0.634 \\
 & \textbf{AV2T-SAM} &\ding{51} & 85.78 & 0.919 & 65.76 & 0.738 \\
 \midrule
 \multirow{2}{*}{\textbf{SAM2}} 
 & \textbf{AV2T-SAM}&\ding{55} & 85.63& 0.920 & 64.47&  0.704\\
 & \textbf{AV2T-SAM}&\ding{51} &\textbf{86.67}&  \textbf{0.924}&  \textbf{69.65} &  \textbf{0.777}\\
\bottomrule
\end{tabular}
}
\caption{Performance comparison across subsets S4 and MS3 for various methods. In the ``Adapter" column, \ding{51} indicates that the model is trained with the adapter module, while \ding{55} denotes that the model is trained without the adapter.}
\label{tab:main}
\end{table}

We compare our model with state-of-the-art methods in Table~\ref{tab:main}. 
Our proposed model achieves new state-of-the-art performance on both subsets of ASVBench. On the S4 dataset, our model outperform the previous state-of-the-art, SAVE~\cite{nguyen2024save}, with an improvement of 1.56 $\mathcal{M}_{\mathcal{J}}$ without introducing extra adapter layers into every image encoder block. Also, on the MS3 dataset, we exceed the previous state-of-the-art, ST-BAVA~\cite{seon2024extending}, without utilizing their proposed temporal correspondence information.

For a more fair comparison, we also run experiments with the model with SAM1\cite{cen20233dsam} backbone, considering all previous SAM-based approaches employ SAM1 as a backbone. Under this setting, our model continues to surpass all existing state-of-the-art methods on the S4 dataset. On the MS3 dataset, it falls marginally short of the best result but still maintains competitive performance.
When using the SAM2~\cite{ravi2024sam2} backbone, many complex components, such as adapters within every block of the image encoder as in SAVE, and a module for temporal understanding as in ST-BAVA, are unnecessary.


Figure~\ref{fig:examples} compares our segmentation results with those of existing models. In the left portion of the figure, we show one example of the results on the S4 dataset. TPAVI \cite{zhou2022avs} captures only part of the guitar, suggesting weak visual object understanding. SAMA-AVS \cite{liu2024annotation} leverages SAM to accurately segment the guitar but struggles to separate it from nearby objects. In contrast, our model fully segments the guitar and distinguishes it from its surroundings, made possible by strong audio-visual correspondence guided by the pretrained text-prompted SAM. A similar pattern emerges in the right portion of the figure, which depicts results on the MS3 dataset. Again, TPAVI exhibits partial segmentation of objects. While SAMA-AVS identifies the correct objects, it fails to separate the keyboard from a person. Our model, however, successfully segments the keyboard and isolates it from other objects. 

These findings underscore our model’s superior ability to correlate audio with its visual region, thanks to our proposed semantically aligned feature and text-prompted SAM.

\subsection{Ablation on Adapters}
We present an ablation study of our framework in Table~\ref{tab:main}. 
In general, methods without adapters perform not as well as those with adapters, and this also applies to our method. 

Interestingly, comparing all the methods that do not use adapters, our proposed method still achieves much better performance than the previous method without adapters. Moreover, compared to existing methods that rely on adapters, our approach delivers comparable results—an outcome not achieved by any prior methods. This highlights the effectiveness of our framework and feature design.


From the result, we can notice that our framework, even without the adapter module, enhances performance dramatically on S4, the dataset which requires shallow audio-visual fusion. However, the model without the adapter module still suffers on the Multi Sound Sources dataset (MS4), the dataset that requires more profound audio-visual. This observation shows that while our approach strengthens audio-visual correspondence, it cannot replace the adapter module. 


\subsection{Ablation on Features}
In this section, we compare the model's performance with different prompt features. As denoted in a Table.~\ref{tab:feature}, we can observe that $\mathbf{\textit{f}_{CLIP} \odot \textit{f}_{CLAP}}$ performs best on both metrics over both datasets. This implies that $\mathbf{\textit{f}_{CLIP} \odot \textit{f}_{CLAP}}$ feature, which captures information from both audio and visual modalities, contains richer information as the prompt than just the CLAP feature, which contains only audio information.

It is worth noticing that the model prompted with CLIP~\cite{radford2021clip} features surpasses the previous state-of-the-art on the S4 dataset, achieving a score of 86.67 on $\mathcal{M}{\mathcal{J}}$. Given that CLIP lacks any audio-related information, this result suggests that a performance of 86.67 on $\mathcal{M}{\mathcal{J}}$ can be attained using only visual cues. This finding highlights a significant vision bias in the dataset, raising concerns about its effectiveness in evaluating true audio-visual understanding.

\begin{table}[]
\centering
\resizebox{0.8\linewidth}{!}{
\begin{tabular}{ccccc}
\toprule
\textbf{Feature} & \multicolumn{2}{c}{\textbf{S4 (V1S)}} & \multicolumn{2}{c}{\textbf{MS3 (V1M)}} \\ 
\cmidrule(lr){2-3} \cmidrule(lr){4-5}
 & $\mathcal{M}_{\mathcal{J}}$ & $\mathcal{M}_{\mathcal{F}}$ & $\mathcal{M}_{\mathcal{J}}$ & $\mathcal{M}_{\mathcal{F}}$ \\ 
\midrule
 CLIP~\cite{radford2021clip} & 86.29 & 0.920 & 64.23 & 0.738 \\
 CLAP~\cite{elizalde2023clap} & 85.67 & 0.915 & 68.15 & 0.743 \\
$\mathbf{\textit{\textbf{f}}_{CLIP} \odot \textit{\textbf{f}}_{CLAP}}$ & \textbf{86.67}&  \textbf{0.924}&  \textbf{69.65} &  \textbf{0.777} \\
\bottomrule
\end{tabular}
}
\caption{Ablation on the prompt features.}
\label{tab:feature}
\end{table}

\section{Conclusion}
In this work, we proposed a novel framework that leverages text modality embeddings and pre-trained text-prompted SAM on the audio-visual segmentation task.
Also, we introduced a new feature that is aware of the intersection of audio and visual modalities for rich semantic information. Comprehensive experiment results demonstrated that our approach performs comparable on the AVS task to state-of-the-arts models. Finally, we demonstrated the vision bias issue of the S4 dataset by surpassing previous state-of-the-art performance without using audio information. 

\section{Acknowledgments}
You Zhang would like to thank the synergistic activities provided by the NRT program on AR/VR funded by NSF grant \#1922591. The GPU resources are supported by NAIRR Pilot Project \#240152, ACCESS \#ELE240019, and NCSA Delta. 


\clearpage

%
%

\par\vskip -2ex 
\bibliographystyle{IEEEbib}
\bibliography{refs_final}

\begin{thebibliography}{10}

\bibitem{chen2021sslhard}
Honglie Chen, Weidi Xie, Triantafyllos Afouras, Arsha Nagrani, Andrea Vedaldi, and Andrew Zisserman,
\newblock ``Localizing visual sounds the hard way,''
\newblock in {\em CVPR}, 2021.

\bibitem{arandjelovic2018sslobjects}
Relja Arandjelovic and Andrew Zisserman,
\newblock ``Objects that sound,''
\newblock in {\em ECCV}, 2018.

\bibitem{zhou2022avs}
Jinxing Zhou, Jianyuan Wang, Jiayi Zhang, Weixuan Sun, Jing Zhang, Stan Birchfield, Dan Guo, Lingpeng Kong, Meng Wang, and Yiran Zhong,
\newblock ``Audio--visual segmentation,''
\newblock in {\em ECCV}, 2022.

\bibitem{gao2024avsegformer}
Shengyi Gao, Zhe Chen, Guo Chen, Wenhai Wang, and Tong Lu,
\newblock ``{AVSegformer}: Audio-visual segmentation with transformer,''
\newblock in {\em AAAI}, 2024.

\bibitem{kirillov2023sam}
Alexander Kirillov, Eric Mintun, Nikhila Ravi, Hanzi Mao, Chloe Rolland, Laura Gustafson, Tete Xiao, Spencer Whitehead, Alexander~C. Berg, Wan-Yen Lo, Piotr Dollar, and Ross Girshick,
\newblock ``Segment anything,''
\newblock in {\em ICCV}, 2023.

\bibitem{ravi2024sam2}
Nikhila Ravi, Valentin Gabeur, Yuan-Ting Hu, Ronghang Hu, Chaitanya Ryali, Tengyu Ma, Haitham Khedr, Roman R{\"a}dle, Chloe Rolland, Laura Gustafson, Eric Mintun, Junting Pan, Kalyan~Vasudev Alwala, Nicolas Carion, Chao-Yuan Wu, Ross Girshick, Piotr Doll{\'a}r, and Christoph Feichtenhofer,
\newblock ``{SAM 2}: Segment anything in images and videos,''
\newblock {\em arXiv:2408.00714}, 2024.

\bibitem{liu2023prompting}
Xiaoxia Liu, Jingyi Wang, Jun Sun, Xiaohan Yuan, Guoliang Dong, Peng Di, Wenhai Wang, and Dongxia Wang,
\newblock ``Prompting frameworks for large language models: A survey,''
\newblock {\em arXiv preprint arXiv:2311.12785}, 2023.

\bibitem{mo2023av}
Shentong Mo and Yapeng Tian,
\newblock ``Av-sam: Segment anything model meets audio-visual localization and segmentation,''
\newblock {\em arXiv:2305.01836}, 2023.

\bibitem{liu2024annotation}
Jinxiang Liu, Yu~Wang, Chen Ju, Chaofan Ma, Ya~Zhang, and Weidi Xie,
\newblock ``Annotation-free audio-visual segmentation,''
\newblock in {\em WACV}, 2024.

\bibitem{radford2021clip}
Alec Radford, Jong~Wook Kim, Chris Hallacy, Aditya Ramesh, Gabriel Goh, Sandhini Agarwal, Girish Sastry, Amanda Askell, Pamela Mishkin, Jack Clark, et~al.,
\newblock ``Learning transferable visual models from natural language supervision,''
\newblock in {\em ICML}, 2021.

\bibitem{elizalde2023clap}
Benjamin Elizalde, Soham Deshmukh, Mahmoud~Al Ismail, and Huaming Wang,
\newblock ``Clap learning audio concepts from natural language supervision,''
\newblock in {\em ICASSP}, 2023.

\bibitem{ma2024medisam}
Jun Ma, Yuting He, Feifei Li, Lin Han, Chenyu You, and Bo~Wang,
\newblock ``Segment anything in medical images,''
\newblock {\em Nat. Commun.}, vol. 15, no. 1, pp. 654, 2024.

\bibitem{lin2024posesam}
Jiehong Lin, Lihua Liu, Dekun Lu, and Kui Jia,
\newblock ``Sam-6d: Segment anything model meets zero-shot 6d object pose estimation,''
\newblock in {\em CVPR}, 2024.

\bibitem{wang2024remotesam}
Di~Wang, Jing Zhang, Bo~Du, Minqiang Xu, Lin Liu, Dacheng Tao, and Liangpei Zhang,
\newblock ``Samrs: Scaling-up remote sensing segmentation dataset with segment anything model,''
\newblock {\em NeurIPS}, 2024.

\bibitem{zhang2024evfsam}
Yuxuan Zhang, Tianheng Cheng, Rui Hu, Lei Liu, Heng Liu, Longjin Ran, Xiaoxin Chen, Wenyu Liu, and Xinggang Wang,
\newblock ``{EVF-SAM}: Early vision-language fusion for text-prompted segment anything model,''
\newblock {\em arXiv:2406.20076}, 2024.

\bibitem{Zhao_2018_pixel}
Hang Zhao, Chuang Gan, Andrew Rouditchenko, Carl Vondrick, Josh McDermott, and Antonio Torralba,
\newblock ``The sound of pixels,''
\newblock in {\em ECCV}, 2018.

\bibitem{Chen_2021_hard}
Honglie Chen, Weidi Xie, Triantafyllos Afouras, Arsha Nagrani, Andrea Vedaldi, and Andrew Zisserman,
\newblock ``Localizing visual sounds the hard way,''
\newblock in {\em CVPR}, 2021.

\bibitem{li2023catr}
Kexin Li, Zongxin Yang, Lei Chen, Yi~Yang, and Jun Xiao,
\newblock ``{CATR}: Combinatorial-dependence audio-queried transformer for audio-visual video segmentation,''
\newblock in {\em ACM MM}, 2023.

\bibitem{liu2023avstwostages}
Chen Liu, Peike~Patrick Li, Xingqun Qi, Hu~Zhang, Lincheng Li, Dadong Wang, and Xin Yu,
\newblock ``Audio-visual segmentation by exploring cross-modal mutual semantics,''
\newblock in {\em Proceedings of the 31st ACM International Conference on Multimedia}, 2023, pp. 7590--7598.

\bibitem{seon2024extending}
Juhyeong Seon, Woobin Im, Sebin Lee, Jumin Lee, and Sung-Eui Yoon,
\newblock ``Extending segment anything model into auditory and temporal dimensions for audio-visual segmentation,''
\newblock in {\em 2024 IEEE International Conference on Image Processing (ICIP)}. IEEE, 2024.

\bibitem{nguyen2024save}
Khanh-Binh Nguyen and Chae~Jung Park,
\newblock ``{SAVE}: Segment audio-visual easy way using segment anything model,''
\newblock {\em arXiv:2407.02004}, 2024.

\bibitem{kazemzadeh2014referitgamedata}
Sahar Kazemzadeh, Vicente Ordonez, Mark Matten, and Tamara Berg,
\newblock ``Referitgame: Referring to objects in photographs of natural scenes,''
\newblock in {\em EMNLP}, 2014.

\bibitem{yu2016refcocodata}
Licheng Yu, Patrick Poirson, Shan Yang, Alexander~C. Berg, and Tamara~L. Berg,
\newblock ``Modeling context in referring expressions,''
\newblock in {\em ECCV}, 2016.

\bibitem{nagaraja2016refcocogdata}
Varun~K. Nagaraja, Vlad~I. Morariu, and Larry~S. Davis,
\newblock ``Modeling context between objects for referring expression understanding,''
\newblock in {\em ECCV}, 2016.

\bibitem{wang2022beit3}
Wenhui Wang, Hangbo Bao, Li~Dong, Johan Bjorck, Zhiliang Peng, Qiang Liu, Kriti Aggarwal, Owais~Khan Mohammed, Saksham Singhal, Subhojit Som, and Furu Wei,
\newblock ``Image as a foreign language: Beit pretraining for vision and vision-language tasks,''
\newblock in {\em CVPR}, 2023.

\bibitem{cen20233dsam}
Jiazhong Cen, Zanwei Zhou, Jiemin Fang, Chen Yang, Wei Shen, Lingxi Xie, Xiaopeng Zhang, and Qi~Tian,
\newblock ``Segment anything in 3d with nerfs,''
\newblock in {\em NeurIPS}, 2023.

\end{thebibliography}

\end{document}